%% file: main.tex
\documentclass[10pt,twocolumn,letterpaper]{article}

\usepackage[final]{cvpr}       

\usepackage[table]{xcolor}
\usepackage{capt-of}
\usepackage{xspace}
\usepackage{xcolor}
\usepackage{enumitem}
\usepackage{amsmath,amsthm,amssymb,amsfonts,dsfont,pifont,bm,bbm,mathrsfs,mathtools,nicefrac,extarrows,relsize}
\usepackage{algorithm,algpseudocode,listings}
\usepackage{booktabs,multirow,adjustbox,diagbox,threeparttable,tabularray,setspace}

\definecolor{cvprblue}{rgb}{0.21,0.49,0.74}
\usepackage[pagebackref,breaklinks,colorlinks,citecolor=cvprblue,bookmarks=false]{hyperref}
\usepackage{wrapfig}
\usepackage[capitalize]{cleveref}  

\crefname{section}{Sec.}{Secs.}
\Crefname{section}{Section}{Sections}
\crefname{appendix}{App.}{Apps.}
\Crefname{appendix}{Appendix}{Appendices}
\crefname{table}{Tab.}{Tabs.}
\Crefname{table}{Table}{Tables}
\crefname{figure}{Fig.}{Figs.}
\Crefname{figure}{Figure}{Figures}
\crefname{equation}{Eq.}{Eqs.}
\Crefname{equation}{Equation}{Equations}
\crefname{theorem}{Thm.}{Thms.}
\Crefname{theorem}{Theorem}{Theorems}
\crefname{lemma}{Lem.}{Lems.}
\Crefname{lemma}{Lemma}{Lemmas}
\crefname{remark}{Rem.}{Rems.}
\Crefname{remark}{Remark}{Remarks}
\crefname{corollary}{Cor.}{Cors.}
\Crefname{corollary}{Corollary}{Corollaries}
\crefname{algorithm}{Alg.}{Algs.}
\Crefname{algorithm}{Algorithm}{Algorithms}
\definecolor{cellred}{RGB}{213, 123, 101}
\definecolor{cellgreen}{RGB}{0, 205, 0}
\definecolor{cellblue}{RGB}{54, 125, 189}
\definecolor{codegreen}{rgb}{0,0.6,0}
\definecolor{codegray}{rgb}{0.5,0.5,0.5}
\definecolor{codepurple}{rgb}{0.58,0,0.82}
\definecolor{backcolour}{rgb}{1.0,1.0,1.0}
\lstdefinestyle{mystyle}{
    backgroundcolor=\color{backcolour},
    commentstyle=\color{codegreen},
    keywordstyle=\color{magenta},
    numberstyle=\tiny\color{codegray},
    stringstyle=\color{codepurple},
    basicstyle=\ttfamily\scriptsize,
    breakatwhitespace=false,
    breaklines=true,
    captionpos=b,
    keepspaces=true,
    numbers=left,
    numbersep=5pt,
    showspaces=false,
    showstringspaces=false,
    showtabs=false,
    tabsize=2
}
\newcommand{\tocite}[1]{{\color{red} [TO CITE]}}
\newcommand{\methodname}{HoloCine}
\newcommand{\method}{\texttt{\methodname}\xspace}
\def\confName{CVPR}
\def\confYear{2026}

\title{HoloCine: Holistic Generation of Cinematic Multi-Shot Long Video Narratives}

\author{Yihao Meng$^{1,2}$ \and
Hao Ouyang$^{2}$ \and
Yue Yu$^{1,2}$ \and
Qiuyu Wang$^{2}$ \and
Wen Wang$^{2,3}$ \and
Ka Leong Cheng$^{2}$ \and
Hanlin Wang$^{1,2}$ \and
Yixuan Li$^{2,4}$ \and
Cheng Chen$^{2,5}$ \and
Yanhong Zeng$^{2}$ \and
Yujun Shen$^{2}$ \and
Huamin Qu$^{1}$ \and \hspace{0.9\linewidth}
\and $^{1}$ HKUST \and $^{2}$ Ant Group \and $^{3}$ ZJU \and $^{4}$ CUHK \and $^{5}$ NTU
}

\begin{document}
\twocolumn[{
    \renewcommand\twocolumn[1][]{#1}
    \maketitle
    \begin{center}
      \vspace{-2pt}
      \includegraphics[width=\textwidth]{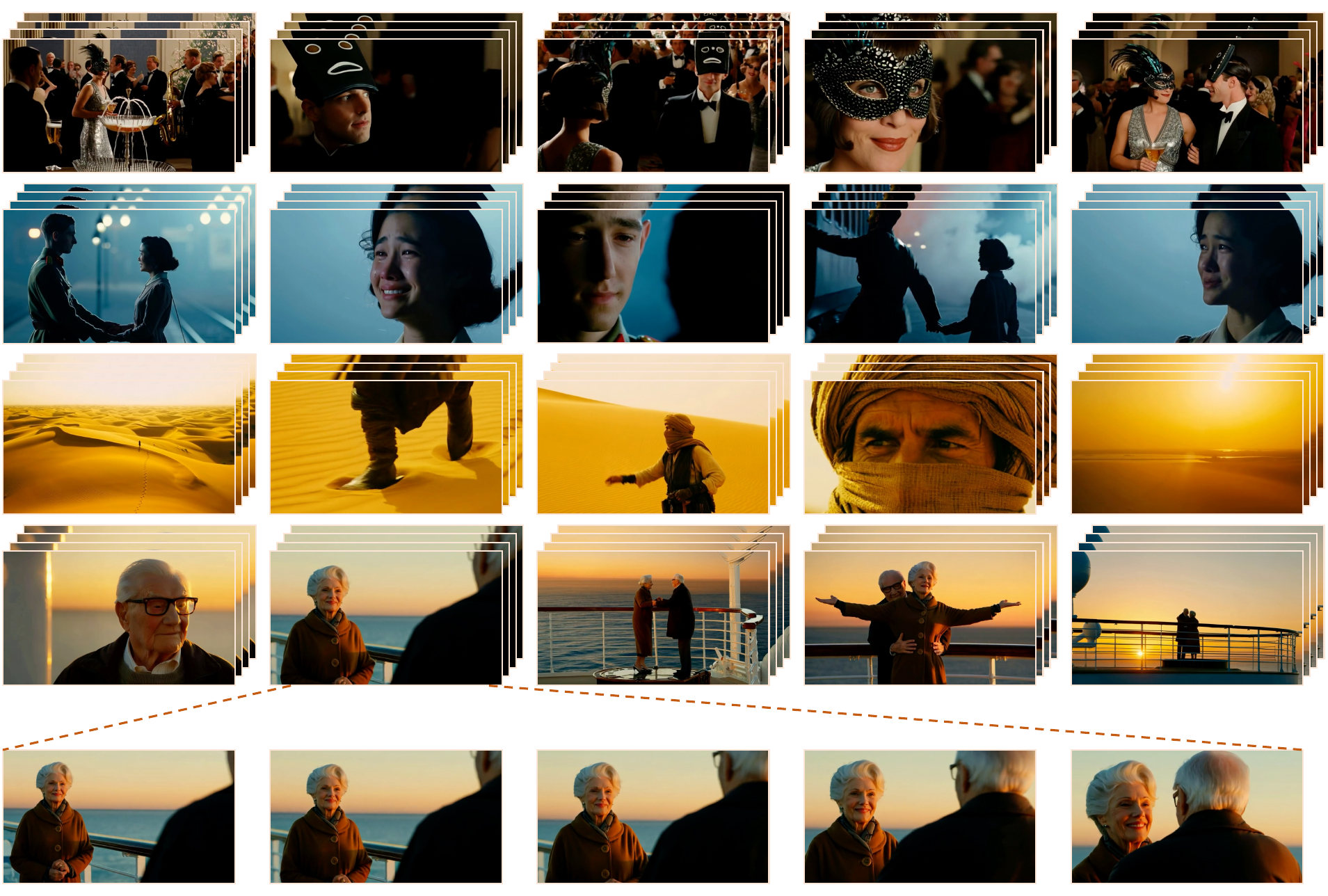}
      \vspace{-20pt}
      \captionsetup{type=figure}
      \caption{From a text prompt alone, \method generates coherent cinematic multi-shot video narratives in a holistic pass. The figure showcases our model's versatility, featuring diverse original scenes (top three rows) and a cinematic homage to Titanic (bottom rows). All scenes exhibit exceptional character consistency and narrative coherence. The expanded final row demonstrates smooth intra-shot motion and quality. Our code is available at:
\textcolor{magenta}{\href{https://holo-cine.github.io/}{https://holo-cine.github.io/}}.}

      \label{fig:teaser}
      \vspace{10pt}
    \end{center}
}]

\maketitle

\input{sections/0_abstract.tex}

\input{sections/1_intro.tex}

\input{sections/2_related.tex}
\input{sections/3_method.tex}
\input{sections/4_exp.tex}
\input{sections/5_conclusion.tex}
\newpage
\input{sections/6_ref.tex}
\input{sections/7_appendix.tex}

\end{document}

%% file: sections/0_abstract.tex
\begin{abstract}

State-of-the-art text-to-video models excel at generating isolated clips but fall short of creating the coherent, multi-shot narratives, which are the essence of storytelling. We bridge this ``narrative gap'' with \method, a model that generates entire scenes holistically to ensure global consistency from the first shot to the last. Our architecture achieves precise directorial control through a Window Cross-Attention mechanism that localizes text prompts to specific shots, while a Sparse Inter-Shot Self-Attention pattern (dense within shots but sparse between them) ensures the efficiency required for minute-scale generation. Beyond setting a new state-of-the-art in narrative coherence, \method develops remarkable emergent abilities: a persistent memory for characters and scenes, and an intuitive grasp of cinematic techniques. Our work marks a pivotal shift from clip synthesis towards automated filmmaking, making end-to-end cinematic creation a tangible future.

\end{abstract}

%% file: sections/1_intro.tex
\section{Introduction}\label{sec:intro}
The field of generative AI has witnessed extraordinary progress, with text-to-video (T2V) synthesis emerging as a prominent frontier. Driven by the scaling of Diffusion Models~\cite{ddpm} and Diffusion Transformers (DiTs)~\cite{dit}, state-of-the-art models~\cite{wan,hunyuan,kling,yang2024cogvideox,sora} can now generate high-fidelity, single-shot video clips from textual prompts. Yet, this capability falls short of emulating the structure of most visual media. Films, television series, and documentaries are not single, unbroken takes. They are narratives constructed from sequences of distinct shots edited together to tell a cohesive story. This disconnect between current generative capabilities and the language of cinema represents the next major challenge: bridging the \textit{``narrative gap''} by moving from single-clip generation to multi-shot, scene-level synthesis.

Current approaches to generating longer multi-shot videos often rely on a decoupled generation paradigm. Whether generating a video chunk-by-chunk~\cite{causvid,framepack,gen-l-video,diffusion-forcing,streamingt2v,consistI2v,longanimation}, or first creating keyframes and then independently synthesizing the connecting shots~\cite{in-context-lora,storydiffusion,moviedreamer,VideoGen-of-Thought,captain-cinema}, these methods model different parts of the video through separate processes. Even when conditioned on character or scene information to improve consistency~\cite{dreambooth,videostudio,concept_master,filmaster}, the generation of individual shots remains largely independent. This fundamental decoupling inherently limits long-range coherence, leading to prevalent issues like error accumulation and consistency drift, where visual attributes such as character identity and background details degrade over time.

A more promising, emerging direction is the holistic pipeline, recently exemplified by LCT~\cite{LCT}, where the entire multi-shot sequence is modeled jointly. This approach is powerful for maintaining global consistency but introduces two formidable challenges. First, achieving precise  control is difficult, as per-shot instructions can be ``diluted'' within the context of the entire prompt. Second, the prohibitive computational cost of the self-attention mechanism, which scales quadratically with sequence length, makes generating longer, minute-scale videos practically intractable.

In this paper, we introduce \method, a novel framework that unlocks the potential of holistic generation through two specialized architectural designs. For precise directorial control, our Window Cross-Attention mechanism localizes attention, directly aligning per-shot text prompts with their corresponding video segments to enable sharp, narrative-driven transitions. To overcome the computational bottleneck, our Sparse Inter-Shot Self-Attention leverages a hybrid pattern: it maintains dense attention within shots for motion continuity while using sparse connections based on compact summaries for efficient communication between shots. This design reduces computational complexity to a near-linear relationship with the number of shots, making minute-scale holistic generation feasible. Finally, to enable the training of our framework, we developed a robust data curation pipeline to build a large-scale, hierarchically annotated dataset of multi-shot scenes.

Extensive experiments validate the effectiveness of our proposed framework. \method significantly outperforms strong baselines across major existing paradigms—including powerful pre-trained models~\cite{wan}, two-stage keyframe-to-video pipelines~\cite{storydiffusion,in-context-lora}, and other holistic approaches~\cite{wu2025cinetrans}. Our method establishes a new state-of-the-art in long-term consistency, narrative fidelity, and precise shot transition control. Ablation studies further confirm the critical roles of our novel components: Window Cross-Attention is essential for achieving fine-grained directorial control, while Sparse Inter-Shot Self-Attention is vital for scalability, delivering quality comparable to full attention at a fraction of the computational cost. Finally, our analysis reveals that \method exhibits remarkable emergent capabilities. These include a persistent memory for characters and scene details across multiple shots and nuanced control over cinematic language, suggesting the model has developed a deeper, implicit understanding of visual storytelling. By enabling minute-scale holistic generation, our work shifts the paradigm from isolated clips to directing entire cinematic scenes, paving the way for automated, end-to-end filmmaking.

%% file: sections/2_related.tex
\section{Related Work}\label{sec:related}
\subsection{Single-Shot Video Generation}
The foundation of our work lies in the rapid progress of single-shot text-to-video (T2V) generation. Early models extended GAN architectures to the video domain \cite{gan1,gan2,gan3,gan4}, but the advent of diffusion models marked a paradigm shift in quality and coherence \cite{videocrafter,DynamiCrafter,svd, yang2024cogvideox,hunyuan,wan}. Foundational models like Imagen Video~\cite{imagen}, Make-A-Video~\cite{make-a-video}, and VDM~\cite{VDM} demonstrated the potential of cascaded diffusion and 3D U-Net architectures to generate short, high-fidelity clips. The introduction of the Diffusion Transformer (DiT) architecture~\cite{dit} further improved scalability and generation quality, operating on latent patches and replacing the U-Net's inductive bias with the powerful self-attention mechanism. Models like \textit{Kling}~\cite{kling} and other open-source efforts~\cite{yang2024cogvideox,hunyuan,wan} have shown that scaling DiT-based architectures leads to remarkable capabilities in generating 5-second-long high-resolution single shot videos. However, these models are fundamentally designed for single shot videos and lack explicit mechanisms to construct a coherent narrative across multiple, distinct shots—the very essence of cinematic storytelling.

\subsection{Multi-Shot and Scene-Level Video Generation}
Bridging the gap from single shots to coherent scenes is an active area of research. One major approach is the hierarchical pipeline~\cite{videostudio,wang2025autostory,moviedreamer,TALC,StoryAgent,DreamFactory}, where an LLM first decomposes a story into sequential prompts, and a T2V model then generates each shot independently. To mitigate inconsistencies, works like VideoStudio~\cite{videostudio} and MovieDreamer~\cite{moviedreamer} add constraints using embeddings or visual tokens, but the isolated nature of shot generation remains a fundamental bottleneck.
Another line of work relies on keyframe-based generation, such as Story Diffusion~\cite{stable_diffusion}, IC-LoRA~\cite{in-context-lora}, and Captain Cinema~\cite{captain-cinema}. These methods first create a sequence of consistent keyframes and then generate the video segments between them. Here, consistency is primarily enforced at the keyframe level, while the video infilling for each shot is still performed in isolation. 
Recognizing these limitations, a more promising paradigm has recently emerged: holistic generation~\cite{LCT,wu2025cinetrans,ShotAdapter}. LCT~\cite{LCT} proposes interleaved positional embeddings within MMDiT~\cite{mmdit} architectures to jointly model all shots in a single, unified diffusion process. This approach inherently enforces global consistency, representing a significant conceptual advance.
Aligning with this emerging paradigm, our framework, \method, generates multi-shot videos holistically for inherent consistency. It further introduces two specialized mechanisms, Window Cross-Attention and Sparse Inter-Shot Self-Attention, to simultaneously provide precise directorial control and computational efficiency for practical scene-level synthesis.
\subsection{Long Video Generation}

Generating long videos, whether single-shot or multi-shot, inevitably confronts the quadratic complexity of the Transformer's self-attention mechanism. The predominant strategy to circumvent this is autoregressive generation, where a video is synthesized in sequential, overlapping chunks~\cite{causvid,framepack,gen-l-video,diffusion-forcing,streamingt2v,consistI2v,longanimation}. To mitigate error accumulation problem in this paradigm, one strategy enhances robustness by training models to denoise controlled noise injected into the historical context (Diffusion Forcing~\cite{diffusion-forcing}). Some methods attempt to distill the entire past into a fixed-size latent vector. For example, TTTVideo~\cite{ttt} encodes context via an MLP at inference time, while FramePack~\cite{framepack} compresses multiple frames into a single vector to predict the next frame. While computationally manageable, this paradigm is prone to consistency drift and error accumulation, where visual fidelity degrades as the sequence lengthens.

To address the computational bottleneck more directly, another line of research explores efficient Transformer architectures. STA~\cite{STA} utilizes localized 3D windows, processing video tile-by-tile in a manner compatible with FlashAttention~\cite{dao2022flashattention}. SageAttention~\cite{zhang2024sageattention} combine selective token compression with a softmax-aware pass to prune key steps of the attention calculation. Radial Attention~\cite{radial_attn} employs a static $O(n \log n)$ mask derived from spatio-temporal energy decay, which enables longer video generation with quality that closely resembles dense attention. Most recent work such as Mixture of Contexts~\cite{moc} has applied similar ideas to long-context video generation by partitioning tokens into chunks and using a trainable router to select relevant context. Our Sparse Inter-Shot Self-Attention is inspired by this line of work but is specifically tailored to the unique structure of multi-shot video. We hypothesize that the information required for consistency between shots is different from that required within a shot, allowing for a principled, structured sparsity pattern that is both efficient and effective for narrative synthesis.

%% file: sections/3_method.tex
\begin{figure*}[t] 
    \centering
    \includegraphics[width=\textwidth]{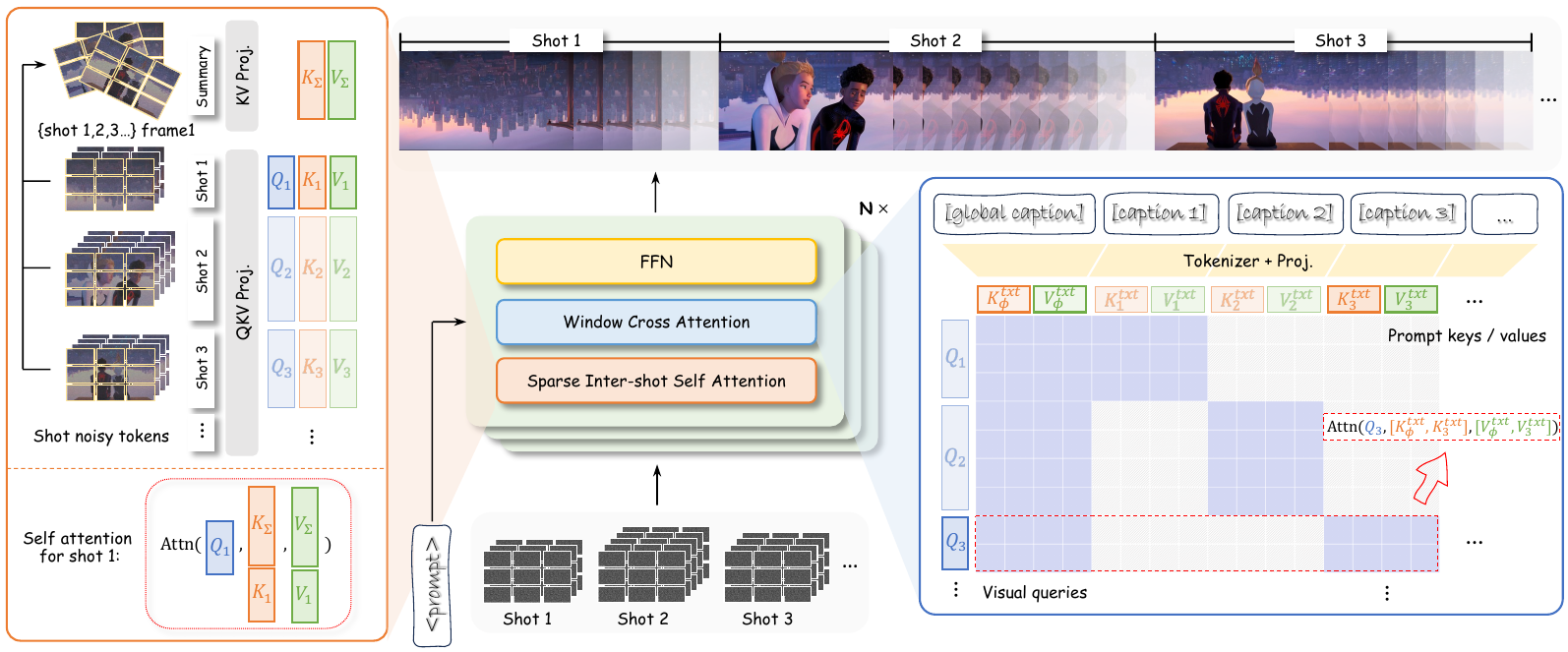} 
    \caption{The architecture of our holistic generation pipeline, where all shot latents are processed jointly. The Window Cross-Attention provides precise directorial control by aligning each shot to its specific text prompt. The Sparse Inter-shot Self-Attention drastically reduces computational cost while preserving long-range consistency. } 
    \vspace{-2ex} 
    \label{fig:overview} 
\end{figure*}

\section{Method}\label{sec:method}

Our goal is to generate a coherent, multi-shot video sequence from a hierarchical text prompt in a single, holistic pass. To achieve this, we introduce \method, a framework built upon the powerful DiT-based video diffusion model, Wan2.2~\cite{wan}.   In the following sections, we detail our data curation and hierarchical annotation pipeline (\cref{sec:data curation}), the Window Cross-Attention mechanism for explicit shot boundary control (\cref{sec:window_cross_attn}), and the Sparse Inter-Shot Self-Attention mechanism that makes holistic generation computationally efficient (\cref{sec:sparse_attn}).

\subsection{Data Curation and Annotation} \label{sec:data curation}
One primary obstacle for multi-shot video generation is the lack of large-scale, high-quality datasets. Public video datasets are typically composed of isolated, short video clips. To address this, we developed a comprehensive data curation pipeline to process cinematic films and television series into a structured, multi-shot dataset.

\noindent\textbf{Shot Segmentation and Filtering.} Our pipeline begins by collecting a large corpus of cinematic content from public sources. We then employ a shot boundary detection algorithm~\cite{transnetV2} to segment each video into individual shots, recording their start and end timestamps. These clips then undergo a rigorous filtering process, where we remove subtitles using~\cite{YaoFANGUK_video-subtitle-extractor_2021} and discard clips that are too short, overly dark, or have low aesthetic scores.

\noindent \textbf{Multi-Shot Sample Assembly:} To construct coherent multi-shot samples, we sequentially group temporally contiguous shots from the source video to form training samples. This grouping is guided by a target total duration (e.g., 5, 15, or 60 seconds), with shots being aggregated until the threshold is met within a certain tolerance. This process generates a diverse set of samples with varying numbers of shots, creating uniform batches for efficient training. The final dataset contains 400k samples with a controllable distribution of shots across these duration tiers.

\noindent \textbf{Hierarchical Captioning:} Each multi-shot sample is annotated with a hierarchical prompt structure using Gemini 2.5 Flash~\cite{gemini_2.5_pro}. A global prompt describes the overarching scene, including the characters, environment, and plot. Following this, a series of per-shot prompts detail the specific actions, camera movements, and characters present in each individual shot~\cite{LCT}. A special \texttt{[shot cut]} tag is inserted between per-shot prompts to explicitly delineate shot boundaries. This two-tier structure provides the model with both global context and fine-grained, temporally localized guidance.

\subsection{Holistic Multi-Shot Generation} \label{sec:holistic_framework}
The foundation of \method is its holistic generation process, where the latent representations for all shots in a video are processed simultaneously within the diffusion model. This joint processing, primarily through a shared self-attention mechanism, allows the model to naturally maintain long-range consistency in aspects like character identity, background, and overall style, ensuring cohesion across all shot boundaries.

While this holistic design is powerful for maintaining consistency, its practical implementation requires careful consideration of two key aspects. First, the model needs explicit guidance to align specific per-shot instructions with their corresponding visual segments. Without a mechanism to localize control, the textual guidance for any given shot would be ``diluted'' by the context of the entire prompt, making it difficult to execute precise control over the per-shot content and shot boundaries. Second, the computational cost of full self-attention, which scales quadratically ($O(L^2)$)  with the sequence length $L$, becomes a prohibitive bottleneck for generating longer, minute-scale videos.

Our architecture directly integrates two specialized mechanisms to address these aspects: Window Cross-Attention for precise directorial control, and Sparse Inter-Shot Self-Attention for computational efficiency.

\subsubsection{Window Cross-Attention} \label{sec:window_cross_attn}
The Window Cross-Attention mechanism is designed to provide precise directorial control, addressing two fundamental requirements simultaneously: \textbf{what} to generate in each shot and \textbf{when} to transition between them. It achieves this by creating a localized link between segments of the video and segments of the text prompt.

Instead of allowing all video tokens to attend to the entire text prompt, our mechanism constrains the attention field to enforce a localized alignment. This is achieved by structuring the attention pattern based on the prompt's hierarchy. For the full sequence of video tokens, the attention each token can pay to the concatenated text prompt is not uniform; rather, it is selectively partitioned. Let $Q_i$ be the query tokens corresponding to the $i$-th shot. We restrict $Q_i$ to only attend to the key-value pairs derived from the global prompt ($KV_{\text{global}}^{\text{txt}}$) and its corresponding $i$-th per-shot prompt ($KV_{i}^{\text{txt}}$). This operation is formally expressed as:
\begin{equation}
\label{eq:window_attention}
\operatorname{Attention}(Q_i, KV^{\text{txt}}) = \operatorname{Attention}\left(Q_i, \left[KV_{\text{global}}^{\text{txt}}, KV_{i}^{\text{txt}}\right]\right)
\end{equation}
This localized attention provides an unambiguous signal for the model to execute sharp, temporally-aligned shot transitions, effectively allowing the text prompt to ``direct'' the shot cuts.





\subsubsection{Sparse Inter-Shot Self-Attention} \label{sec:sparse_attn}
While the holistic design enables high-quality generation, applying full self-attention across the entire sequence of video tokens is computationally prohibitive for longer videos. To overcome this, we propose a Sparse Inter-Shot Self-Attention mechanism, which drastically reduces complexity while preserving necessary information flow.

Our key intuition is that the nature of consistency differs within a shot versus between shots. Specifically, \textbf{intra-shot consistency} demands dense, frame-to-frame temporal modeling to ensure smooth motion and action continuity. In contrast, \textbf{inter-shot consistency} is primarily concerning the persistence of characters, environment, and style—which does not require every frame of one shot to attend to every frame of another.
Based on this, we structure our self-attention as follows. 

\noindent\textbf{Intra-Shot Attention:} Within each shot $i$, we perform full, bidirectional self-attention. The query tokens $Q_i$ from shot $i$ attend to all key-value pairs $KV_i$ from the same shot.

\noindent\textbf{Inter-Shot Attention:} To facilitate information exchange between shots, we create a global context summary. For each shot $j$, we select a small, representative subset of its key-value tokens, $KV_{\text{summary},j}$ (e.g., tokens from the first frame of this shot). These summary tokens from all shots are concatenated to form a global key-value cache, $KV_{\text{global}} = [KV_{\text{summary},1}, \dots, KV_{\text{summary},N_{\text{shots}}}]$. The query tokens $Q_i$ from shot $i$ also attend to this global cache.

The complete self-attention for shot $i$ is formulated as:
\begin{equation}
\label{eq:sparse_attention}
\operatorname{Attention}(Q_i, KV) = \operatorname{Attention}\left(Q_i, \left[KV_{\text{global}}, KV_{i},\right]\right)
\end{equation}

This design drastically reduces computational complexity. If a video has $N_s$ shots of length $L_{\text{shot}}$, and each shot is summarized by $S$ tokens, the total complexity of full attention would be $O((N_s L_{\text{shot}})^2)$. Our sparse attention, however, reduces this to approximately $O(N_s \times (L_{\text{shot}}^2 + L_{\text{shot}} \cdot N_s \cdot S))$. Since $S$ (e.g., the number of tokens in one frame) is much smaller than $L_{\text{shot}}$, this complexity is significantly lower and scales much more favorably with the number of shots, making it feasible to holistically generate minute-level and longer multi-shot videos. We conduct ablation studies on the method for selecting summary tokens, such as using the first frame, first and last frames, or a learnable mechanism.

%% file: sections/4_exp.tex
\section{Experiments}\label{sec:exp}
\begin{figure*}[t] 
    \centering
    \includegraphics[width=\textwidth]{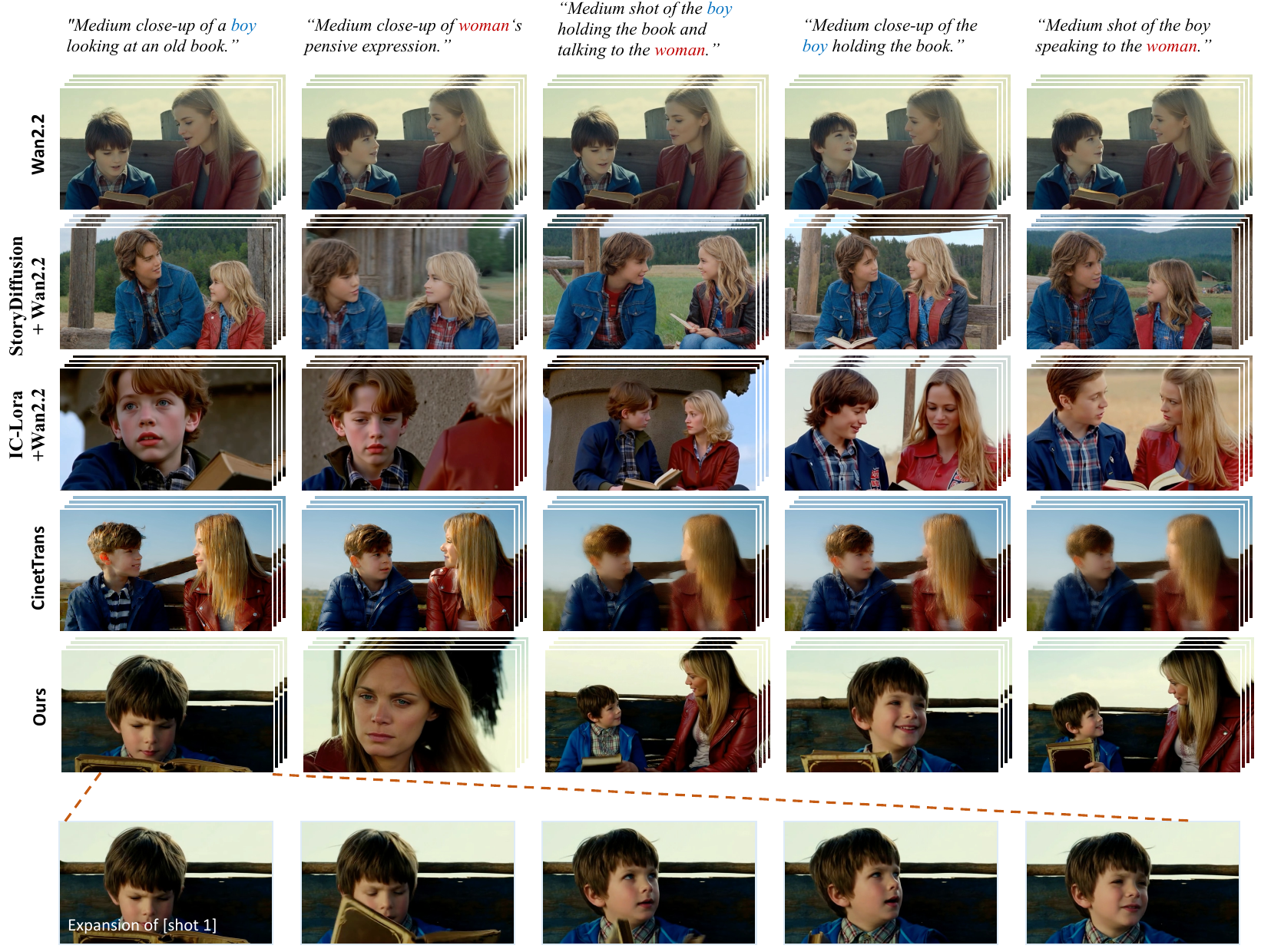} 
    \caption{Qualitative comparison on a complex multi-shot prompt. Our method successfully generates a coherent sequence of distinct shots aligned with per-shot descriptions, while baseline methods fail in maintaining consistency, prompt fidelity, or handling shot transitions.} 
    \vspace{-2ex} 
    \label{fig:comparison} 
\end{figure*}

In this section, we present a comprehensive experimental evaluation of our proposed framework, \method. ~\cref{ssec:implementation} describes the training and implementation details of \method. ~\cref{ssec:comparison_setup} introduce the baselines and metrics for the cinematic multi-shot video generation task, and demonstrate our superior performance over these baselines. In Section~\cref{ssec:ablation}, we analyze the effect of our key proposed modules, including the Window Cross-Attention and Sparse Inter-Shot Self-Attention mechanisms.
In~\cref{ssec:emerging}, we discuss some advanced capabilities of our model, including emergent memory capability. and controllability of cinematographic language.
\subsection{Implementation Details}
\label{ssec:implementation}

\textbf{Training Setup.} Our framework is built upon the 14B parameter version of \texttt{wan2.2}, a powerful DiT-based video diffusion model, which we adapt for the multi-shot task. We train our model on our curated dataset of 400k multi-shot video samples. The dataset includes videos at multiple duration levels (5s, 15s, and 60s) with a maximum of 13 shots per video, and all samples are processed at a resolution of $480 \times832$. The model is trained for 10k steps with a learning rate of $1 \times 10^{-5}$ and a linear warmup schedule. The entire training process is conducted on 128 NVIDIA H800 GPUs. To manage the significant memory requirements of training on such long video sequences, we employ a hybrid parallelism strategy, using Fully Sharded Data Parallelism (FSDP) to shard the model parameters and Context Parallelism (CP) to split the long token sequences across multiple devices.

\noindent \textbf{Attention Implementation.} The implementation of our proposed attention mechanisms is optimized for efficiency. For our Sparse Inter-Shot Self-Attention, where computational cost is a primary concern, we leverage the highly efficient \texttt{varlen} (variable-length) sequence functionality from FlashAttention-3\cite{dao2022flashattention}. For each query shot, we construct its corresponding Key and Value context by concatenating its own dense local tokens with the shared global summary tokens. These resulting variable-length sequences are then packed into single tensors, which allows the GPU to compute the complex, sparse attention pattern in a single, optimized kernel launch without any overhead from padding tokens. In contrast, for the Window Cross-Attention, since the text prompt sequences are short and this operation constitutes a small fraction of the total computation, we simply apply an attention mask to restrict the attention region. This approach is highly effective and incurs negligible performance overhead.


\subsection{Comparison}
\label{ssec:comparison_setup}

\noindent \textbf{Settings.}
We compare \method against three categories of strong baselines representing the main paradigms for multi-shot long video generation:

\begin{itemize}
    \item \textbf{Pre-trained Video Diffusion Model.} We test the capability of the powerful pre-trained video diffusion model, \texttt{Wan2.2 14B}~\cite{wan}, for the multi-shot task. We provide the model with our full hierarchical prompt (concatenated global and per-shot descriptions) and task it with generating the entire multi-shot sequence in one pass. This baseline evaluates whether a state-of-the-art model can understand and execute multi-shot instructions without our proposed architectural modifications.

\item \textbf{Two-Stage Keyframe-to-Video Generation.} This paradigm first generates a set of consistent keyframes, one for each shot, and then uses a powerful I2V model to animate them into video clips. We evaluate two state-of-the-art methods for the keyframe generation stage: StoryDiffusion~\cite{storydiffusion}, which produces a complete multi-shot image sequence, and IC-LoRA~\cite{in-context-lora}, which generates keyframes using in-context learning. For a fair comparison, we utilize our base model, the \texttt{wan2.2 14B}, as the I2V component for both pipelines.

    \item \textbf{Holistic Multi-Shot Generation.} We compare against CineTrans~\cite{wu2025cinetrans}, a most recent work that also performs holistic generation of multi-shot videos.
\end{itemize}

To facilitate a comprehensive evaluation of the multi-shot video generation task, we constructed a new benchmark dataset. We leveraged the capabilities of Gemini 2.5 Pro~\cite{gemini_2.5_pro} to generate 100 diverse hierarchical text prompts, each containing explicit instructions for shot transitions. This test set spans a wide range of genres and narrative structures, enabling a robust assessment of a model's ability to maintain consistency and control across complex sequences. To ensure a fair comparison, we adapted the hierarchical prompts for the two-stage methods. We generated a distinct prompt for each shot by merging the global context with the shot-specific instructions. This process involved resolving abstract character ID tags (e.g., [character1]) into their full textual descriptions, ensuring all methods received equivalent semantic information.

We note that most related works LCT~\cite{LCT}, Mixture of Concept~\cite{moc}, and Captain Cinema~\cite{captain-cinema} are not open-sourced. Therefore, a direct quantitative comparison is not feasible. We will provide a qualitative comparison against their published results in the appendix.

\noindent \textbf{Evaluation Metrics.}
We evaluate the models on five crucial aspects: overall video quality, semantic consistency(prompt adherence), intra-shot consistency, inter-shot consistency, and transition control. 
For overall quality, prompt adherence, and intra-shot consistency, we utilize the comprehensive VBench~\cite{vbench} benchmark. 
To specifically assess inter-shot consistency, we compute a ViCLIP-based similarity score between pairs of shots that are annotated to contain the same character.  
Furthermore, to better evaluate the model's ability to follow explicit shot-cut instructions, we propose the Shot Cut Accuracy (SCA) metrics.
More details on these evaluation metrics are presented in~\cref{sec:app_metrics}.   
This metric holistically assesses shot control by quantifying both the accuracy of the number of cuts and the temporal precision of their placement.

\label{ssec:results}

\input{tables/quant_compare}

\noindent \textbf{{Quantitative Results.}}
As shown in~\cref{tab:evaluation}, our model \method establishes a new state-of-the-art by achieving superior performance on the vast majority of metrics. It achieves the top scores across all categories central to the multi-shot task: Transition Control, Inter-shot Consistency, Intra-shot Consistency, and Semantic Consistency. While we note that StoryDiffusion+Wan2.2 performs slightly better on Aesthetic Quality, we argue that our holistic generation method, which produces all shots within a unified modeling process, is fundamentally better suited for this task. This architectural choice is precisely why \method excels at maintaining consistency and control, validating its effectiveness in creating coherent narratives where prior paradigms have struggled.

\noindent \textbf{Qualitative Results.}
In~\cref{fig:comparison}, we provide a qualitative comparison on a complex narrative prompt to illustrate the superiority of our method. The pre-trained base model, Wan2.2, fails to comprehend the multi-shot instructions, producing only a single, static shot without any transitions. The two-stage methods, while capable of generating different images, struggle with both prompt fidelity and long-range consistency. For example, the prompt for the second shot is ``Medium close-up of woman's pensive expression,'' yet both StoryDiffusion + Wan2.2 and IC-LoRA + Wan2.2 generate a medium shot of the boy and woman together. Their struggle with long-range consistency is especially evident in shots 4 and 5, where the characters' features diverge significantly from the initial shots. The complexity of the prompt and the required length of the video also prove challenging for CineTrans, causing significant image degradation and preventing it from correctly performing the specified shot transitions. In contrast, our method successfully parses the hierarchical prompt to generate a coherent sequence of five distinct shots. As shown, each shot precisely matches its corresponding text description while maintaining high character and stylistic consistency throughout the entire video, demonstrating the effectiveness of our holistic generation approach.

\begin{figure*}[t] 
    \centering
    \includegraphics[width=\textwidth]{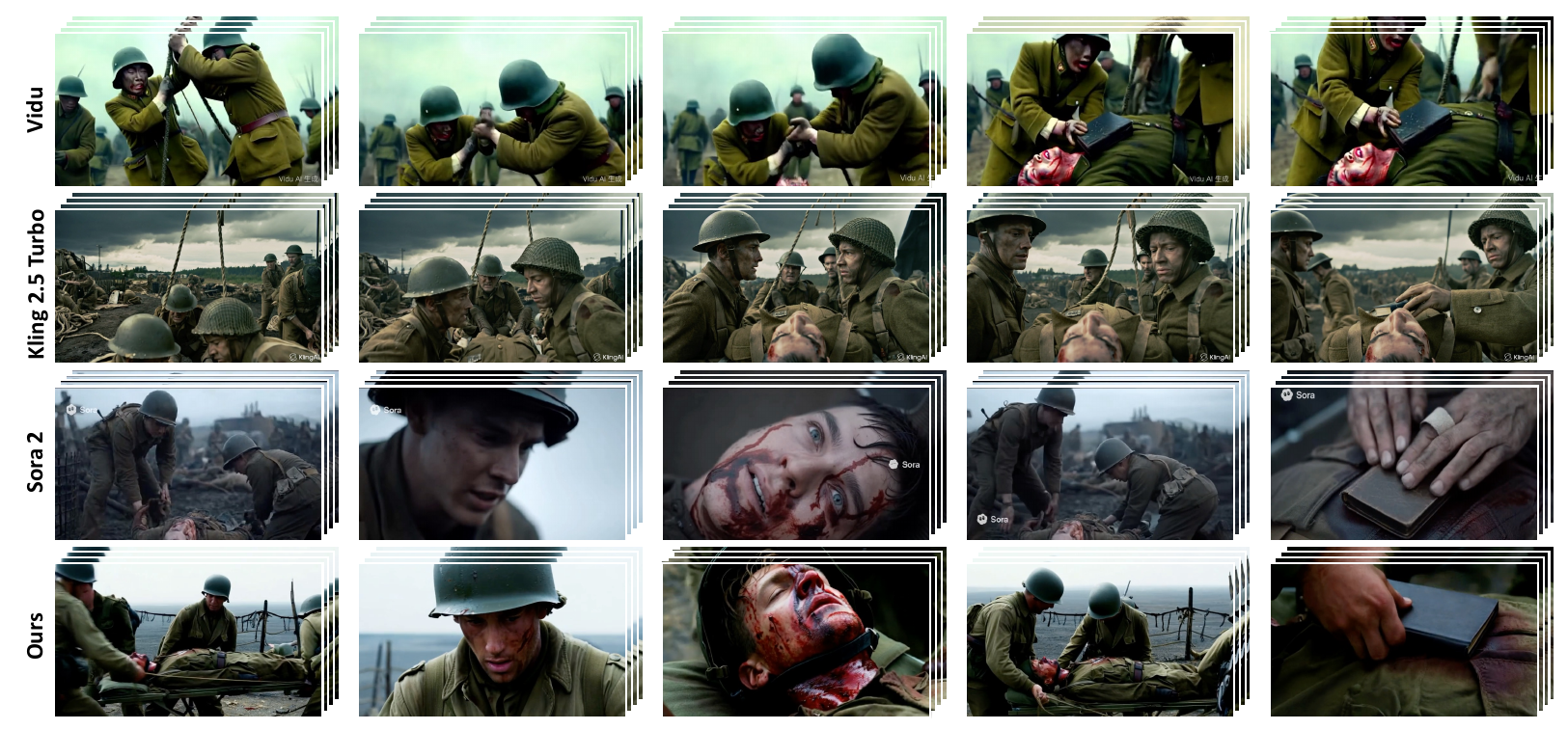} 
    \caption{Qualitative comparison with state-of-the-art commercial models. While Vidu and Kling 2.5 Turbo fail to interpret multi-shot instructions and generate only a single, continuous clip, \method successfully executes complex shot transitions. Our method demonstrates narrative control and consistency comparable to the leading closed-source model, Sora 2, accurately rendering the sequence from medium shots to close-ups as directed by the prompt.} 
    \vspace{-2ex} 
    \label{fig:commercial} 
\end{figure*}

\noindent \textbf{Comparison with Commercial Models.} To further situate \method's capabilities, we conducted a qualitative comparison with leading closed-source commercial models. As illustrated in~\cref{fig:commercial}, while models like Vidu~\cite{vidu2024} and Kling 2.5 Turbo~\cite{kling} generate visually impressive clips, they struggle with the core task of multi-shot storytelling. Given a hierarchical prompt, they produce a single, continuous shot, failing to understand or execute the specified shot transitions. In contrast, \method demonstrates narrative comprehension and control on par with the latest state-of-the-art model, Sora 2~\cite{sora2_2025}. Both models successfully parse the prompt to generate a coherent sequence of distinct shots—transitioning from a medium shot to a dramatic close-up—while maintaining high character and stylistic consistency. This result validates that our framework's ability to create complex, directed narratives is comparable to the leading proprietary solutions in the field.

\subsection{Ablation Studies}

\label{ssec:ablation}
\begin{figure}[ht]
    \centering
    \includegraphics[width=\columnwidth]{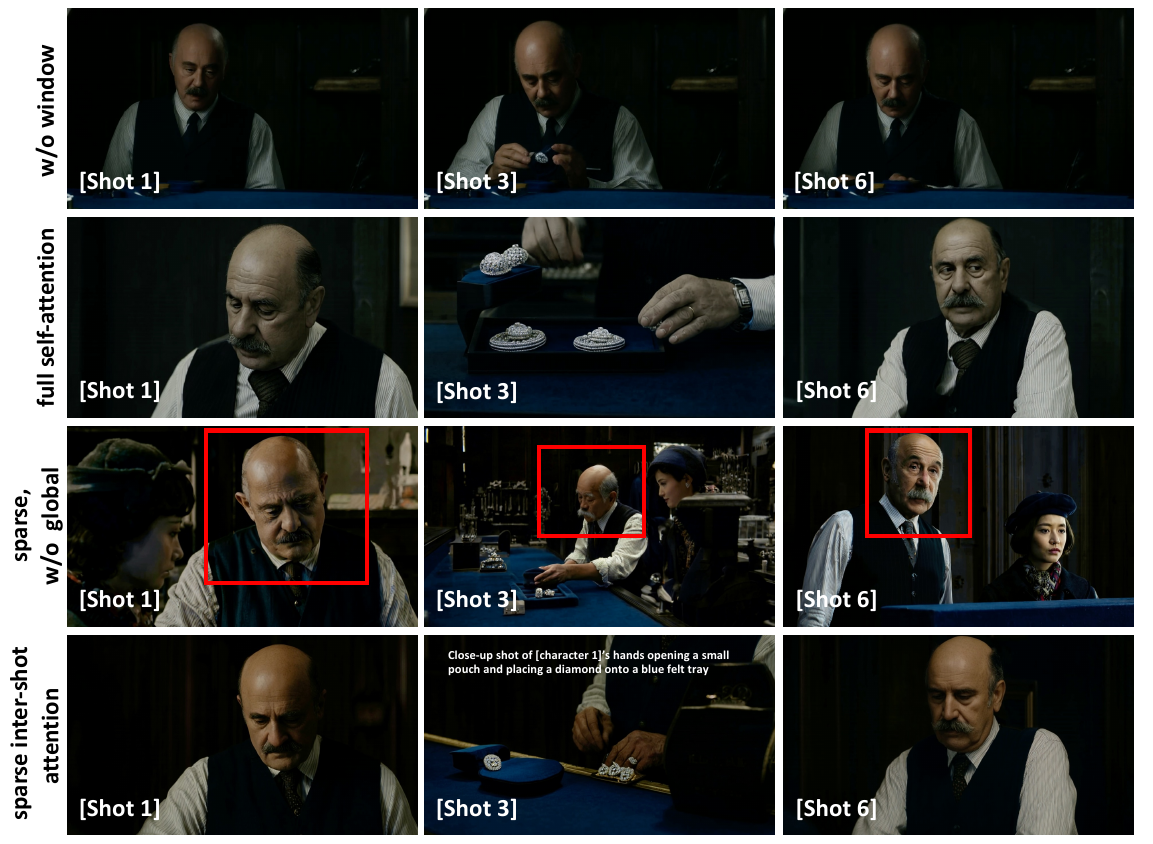}
    \caption{Qualitative results for our ablation study. We compare our full model (bottom row) against three variants. From top to bottom: removing window cross-attention prevents shot transitions; a full self-attention baseline works well but is computationally expensive; and removing inter-shot summary tokens leads to a complete loss of character consistency.}
    \label{fig:ablation}
    \vspace{-10pt}
\end{figure}

\input{tables/ablations}

We perform a series of ablation studies to validate our key architectural choices. The qualitative results are presented in~\cref{fig:ablation}. To facilitate rapid experimentation, all ablation studies were conducted on the wan2.2 5B model.

\noindent \textbf{Window Cross-Attention.} Without our window cross-attention, this model exhibited a severe degradation in shot control, as evidenced by a significantly lower Shot Cut Accuracy (SCA) and per-shot semantic consistency score. As illustrated in the top row of~\cref{fig:ablation},  the model fails to execute shot cuts, ignoring prompt instructions for new content (e.g., the close-up in Shot 3) and remaining locked into the initial scene. This confirms that our windowed attention is crucial for precise shot boundary and content control.

\noindent \textbf{Sparse vs. Full Self-Attention.} We then compare our sparse self-attention to the full, dense attention baseline. While both produce high-quality, consistent videos (second and fourth rows in~\cref{fig:ablation}), the full attention model is computationally prohibitive for generating long sequences. Our sparse attention mechanism, in contrast, provides a highly effective trade-off. It retains the vast majority of the generative quality while offering a fundamental improvement in efficiency and scalability, making complex, scene-level generation feasible.

\noindent \textbf{Inter-Shot Summary Token.} A critical aspect of our sparse attention design is the inter-shot communication facilitated by summary tokens, where each shot attends to the first-frame token of all other shots. To ablate this, we trained a variant where self-attention is confined strictly within each shot, disabling this information exchange. This results in a catastrophic loss of consistency (third row in~\cref{fig:ablation}), where the old man's identity and appearance change drastically between shots. This  demonstrates that our inter-shot summary token mechanism is the critical component for maintaining narrative continuity and character consistency across the entire scene.

\subsection{Advanced Capabilities}
\label{ssec:emerging}

\subsubsection{Emergent Memory Capability}
Beyond generating high-quality and coherent shots, our model exhibits surprising emergent memory. This capability suggests the model is not merely learning superficial visual transitions but is building an implicit and persistent representation of scenes and objects. We demonstrate this memory in three key aspects.

\begin{figure}[ht]
    \centering
    \includegraphics[width=\columnwidth]{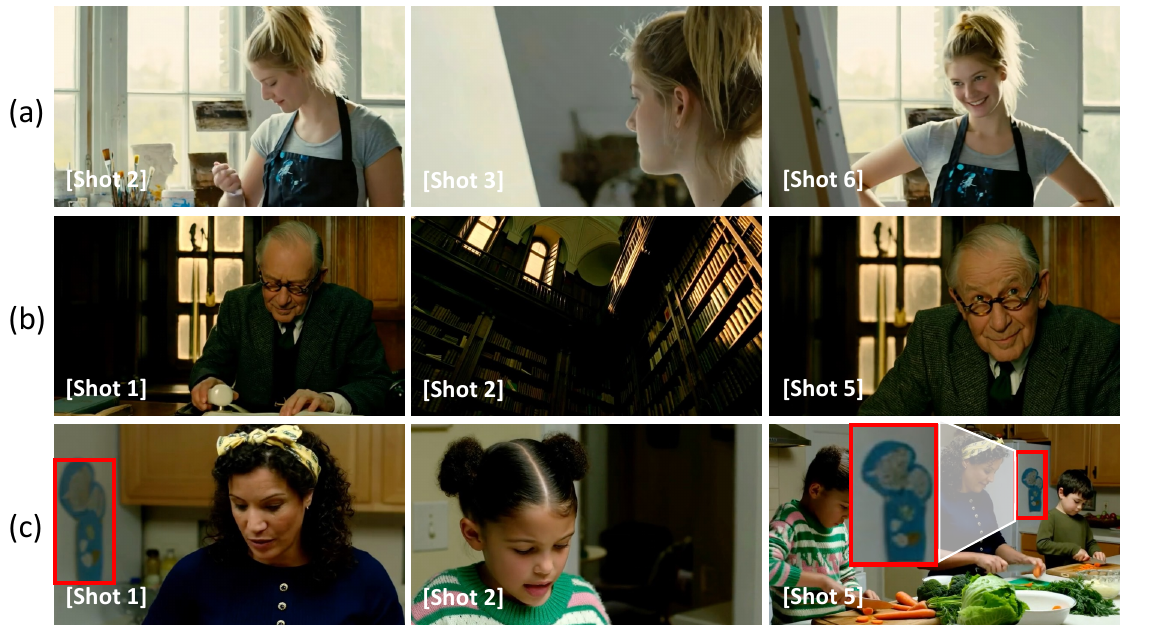}
    \caption{Qualitative results of our model's emergent memory capability. (a) Character Permanence: The subject's identity and appearance are consistently maintained across different camera angles and expressions. (b) Long-range Consistency: The subject is accurately recalled after an unrelated distractor shot. (c) Fine-grained Persistence: A non-salient background detail (the magnet, highlighted) is correctly preserved across an intervening shot.}
    \label{fig:consistency}
    \vspace{-10pt}
\end{figure}

\noindent \textbf{Object/Character Permanence Across Viewpoints.}
Our model maintains consistent character identity across varying shots and angles. In~\cref{fig:consistency}(a), for instance, the artist's key features — her blonde hair, grey t-shirt, and apron—remain identical across a medium shot [Shot 2], a profile view [Shot 3], and a subsequent smiling shot [Shot 6], demonstrating a stable character representation.

\noindent \textbf{Long-range Consistency and Re-appearance.}
The model demonstrates robust long-range consistency, recalling subjects after being interrupted by completely different shots. ~\cref{fig:consistency}(b) shows an A-B-A sequence where a professor, introduced in [Shot 1], is accurately regenerated in [Shot 5] after a distractor shot of the library environment [Shot 2]. His distinct appearance is perfectly preserved, proving a memory that extends beyond adjacent shots.

\noindent \textbf{Fine-grained Detail Persistence.}
Crucially, the model's memory extends to fine-grained, non-salient details, indicating a holistic scene understanding. As illustrated in ~\cref{fig:consistency}(c), a specific blue magnet (highlighted) appears in the background of [Shot 1]. After an intervening shot, the model correctly recalls and renders the exact same magnet in its original position in [Shot 5], despite it not being a central element of the prompt.

\subsubsection{Controllability of Cinematographic Language}
By training on a vast corpus of cinematic data and high-level descriptive prompts, our model has developed a nuanced understanding of filmmaking techniques. Consequently, it exhibits high fidelity in interpreting and executing standard cinematographic commands, enabling precise narrative and stylistic control.

\begin{figure}[ht]
    \centering
    \includegraphics[width=\columnwidth]{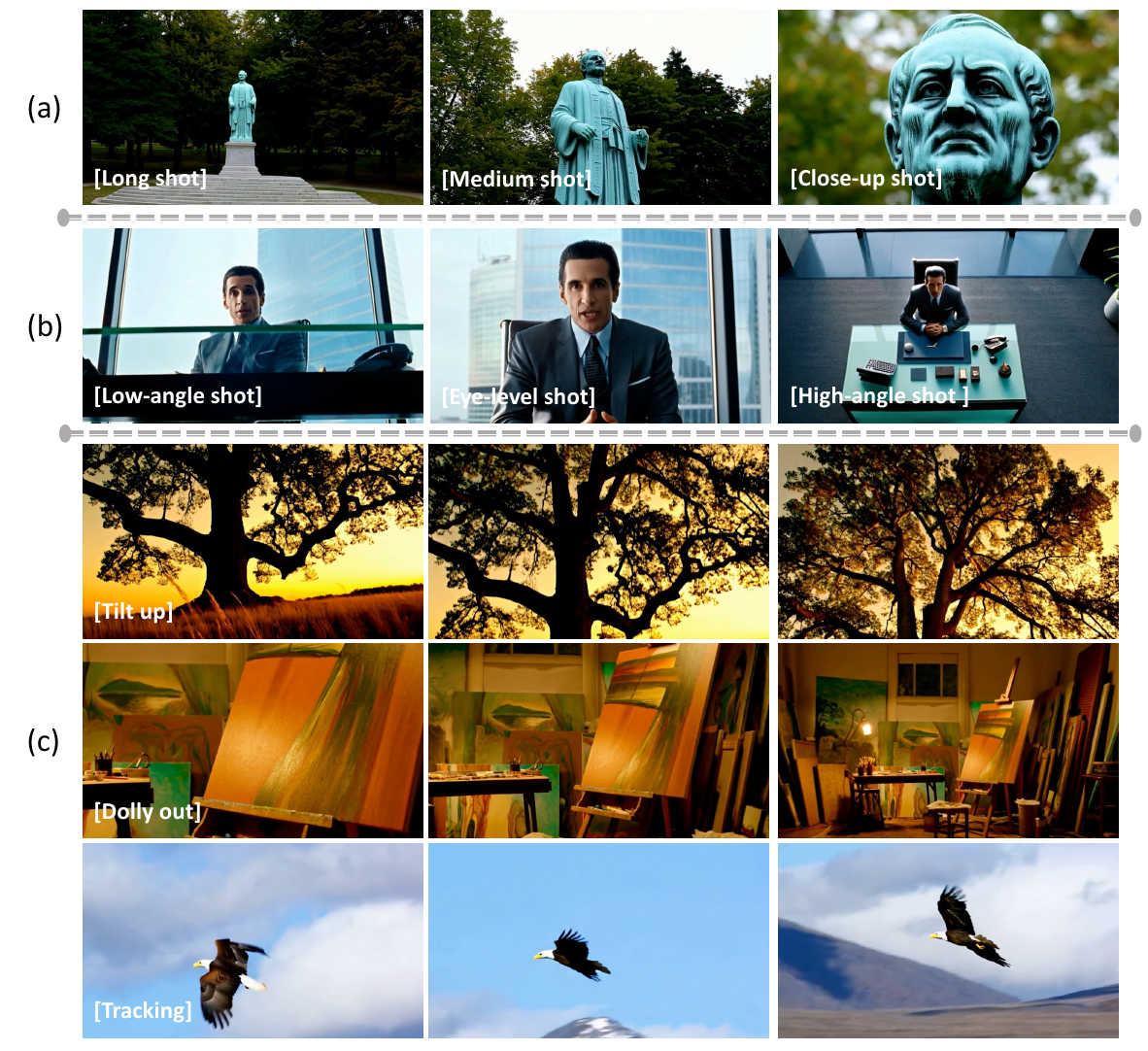}
    \caption{Controllability of Cinematographic Language. Our model demonstrates high fidelity in executing specific cinematic commands. (a) Shot Scale: The model accurately generates long, medium, and close-up shots. (b) Camera Angle: It correctly interprets low-angle, eye-level, and high-angle commands to set the camera's viewpoint. (c) Camera Movement: The model produces fluid and precise camera movements as prompted, including tilt up, dolly out, and tracking shots.}
    \label{fig:camera}
    \vspace{-10pt}
\end{figure}

\noindent \textbf{Shot Scale Control.}
The model accurately renders standard shot scales. As shown in~\cref{fig:camera}(a), given prompts for a [Long shot], [Medium shot], and [Close-up shot] of the same statue, the model generates outputs that correctly correspond to established cinematographic definitions.

\noindent \textbf{Camera Angle Control.}
Our model precisely follows the camera angle commands specified in the text prompt. As shown in~\cref{fig:camera}(b), when prompted with [Low-angle], [Eye-level], and [High-angle] descriptions for the same subject, the model generates the corresponding views accurately. This demonstrates its ability to translate textual cinematographic instructions into correct geometric camera placements within the scene.

\noindent \textbf{Camera Movement Control.}
Our model is capable of producing a wide range of dynamic and fluid camera movements specified in the prompt. As demonstrated in~\cref{fig:camera}(c), the model accurately executes these commands to create compelling visual narratives. For instance, a [Tilt up] command generates a smooth vertical camera motion, gracefully revealing the full height of the tree. A [Dolly out] command results in the camera physically moving backward, progressively unveiling the broader context of the artist's studio. Furthermore, a [Tracking] shot correctly follows the motion of a subject, in this case, keeping the soaring eagle centered in the frame. This mastery over camera movement is crucial for creating professional and engaging cinema sequences.

\subsection{Limitations}

\begin{figure}[ht]
    \centering
    \includegraphics[width=\columnwidth]{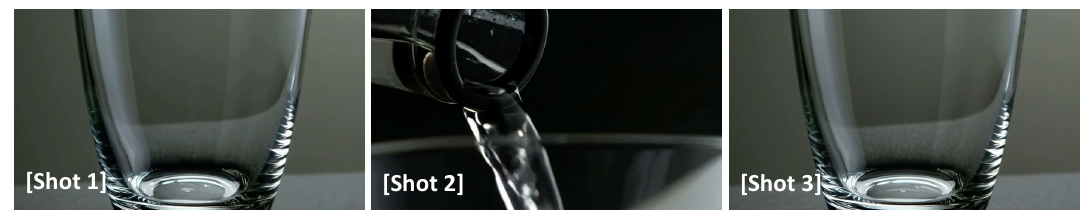}
    \caption{A failure case in causal reasoning. After an action (pouring water, [Shot 2]) is applied to an object (empty glass, [Shot 1]), the model fails to render its logical consequence. It incorrectly reverts to the initial empty state in [Shot 3], prioritizing visual consistency over the action's outcome.}
    \label{fig:limitation}
    \vspace{-10pt}
\end{figure}
While our model excels at maintaining visual consistency, it exhibits limitations in causal reasoning. It can fail to comprehend how an action should alter an object's physical state. ~\cref{fig:limitation} illustrates this clearly. Given an empty glass [Shot 1] and the action of water being poured into it [Shot 2], the model fails to render the logical outcome. Instead, it regenerates the glass as empty in [Shot 3], prioritizing visual consistency with the initial shot over the physical consequence of the action. This highlights a key challenge for future work: advancing from perceptual consistency to logical, cause-and-effect reasoning.

%% file: tables/quant_compare.tex
\begin{table*}[tb]
  \scriptsize
  \caption{\textbf{Quantitative results}. The best and runner-up are in \textbf{bold} and \underline{underlined}.}
  \label{tab:evaluation}
  \centering
  \resizebox{\textwidth}{!}{%
  \renewcommand{\arraystretch}{1.2}%
  \begin{tabular}{p{2.3cm}ccccccccc} %

    \toprule
\multirow{2}{*}{Method} & \multirow{2}{*}{\parbox{1.7cm}{\centering \textbf{Transition Control}\textuparrow}}
  & \multirow{2}{*}{\parbox{1.7cm}{\centering \textbf{Inter-shot Consistency}\textuparrow}} 
  & \multicolumn{2}{c}{\textbf{Intra-shot Consistency}}
  & \multirow{2}{*}{\parbox{1.5cm}{\centering \textbf{Aesthetic Quality}\textuparrow}}
  & \multicolumn{2}{c}{\centering \textbf{Semantic Consistency}} \\
  \cline{4-5} \cline{7-8} %
  & & & \textbf{Subject\textuparrow} & \textbf{Background\textuparrow} & & \textbf{Global\textuparrow} & \textbf{Shot\textuparrow} \\ %

\hline
    %
    Wan2.2 & 0.4843  & 0.6772 & 0.9054 & 0.9014 & 0.5568 & 0.1652 & 0.1364 \\
    CineTrans & \underline{0.5370}  & 0.6152 & 0.8990 & 0.8998 & 0.4789 & 0.1568 & 0.1159 \\
    IC-LoRA+Wan2.2 & -  & 0.7096 & \underline{0.9421} & \underline{0.9303} & 0.5246 & \underline{0.1808} & \underline{0.1692} \\
    StoryDiffusion+Wan2.2 & -  & \underline{0.7364} & 0.8487 & 0.8927 & \textbf{0.5773} & 0.1453 & 0.1644 \\
    \hline
    \methodname (Ours) & \textbf{0.9837}  & \textbf{0.7509} & \textbf{0.9448} & \textbf{0.9352} & \underline{0.5598} & \textbf{0.1856} & \textbf{0.1837} \\
   \bottomrule
  \end{tabular}%
}
\end{table*}

%% file: tables/ablations.tex
\begin{table}[tb]
  \scriptsize
  \caption{\textbf{Ablations}. The best is in \textbf{bold}.}
  \label{tab:ablations}
  \centering
  \resizebox{0.5\textwidth}{!}{%
  \renewcommand{\arraystretch}{1.2}%
  \begin{tabular}{p{2.3cm}cccc} %

    \toprule

    \multirow{2}{*}{Method} & \multirow{2}{*}{\parbox{1.7cm}{\centering \textbf{Transition Control}\textuparrow}}
    & \multirow{2}{*}{\parbox{1.7cm}{\centering \textbf{Inter-shot Consistency}\textuparrow}}
    & \multirow{2}{*}{\parbox{1.5cm}{\centering \textbf{Aesthetic Quality}\textuparrow}}
    & \multirow{2}{*}{\parbox{1.7cm}{\centering \textbf{Semantic Consistency}\textuparrow}} \\
    \\ 

\hline
 
    WO WINDOW & 0.6266  & 0.7009 & \textbf{0.5755} & 0.1562 \\
    FULL ATT WINDOW & 0.8923  & \textbf{0.7231} & 0.5700 & 0.1738 \\
    SPARSE ZERO & 0.9675 & 0.6761 & 0.5669 & 0.1642 \\
    SPARSE & \textbf{0.9736} & 0.7225 & 0.5693 & \textbf{0.1739} \\
   \bottomrule
  \end{tabular}%
}
\end{table}

%% file: sections/5_conclusion.tex
\section{Conclusion}\label{sec:conclusion}
In this work, we bridge the \textit{``narrative gap''} in text-to-video generation with \method, a holistic framework that synthesizes entire multi-shot scenes to ensure global narrative coherence. Our architecture achieves precise directorial control through a Window Cross-Attention mechanism while overcoming prohibitive computational costs with a Sparse Inter-Shot Self-Attention, making minute-scale generation feasible. \method not only establishes a new state-of-the-art in consistency and shot control but also develops remarkable emergent capabilities, such as persistent memory for characters and a nuanced understanding of cinematic language. While our work identifies causal reasoning as a key challenge for future research, \method represents a critical step towards the automated creation of complex visual narratives. By enabling minute-scale holistic generation, it shifts the paradigm from isolated clips to directing entire scenes, making end-to-end film generation a tangible and exciting future.


%% file: sections/6_ref.tex
{
\small
\bibliographystyle{ieeenat_fullname}
\bibliography{main.bbl}
}

%% file: sections/7_appendix.tex
\clearpage
\appendix
\renewcommand\thesection{\Alph{section}}
\renewcommand\thefigure{S\arabic{figure}}
\renewcommand\thetable{S\arabic{table}}
\renewcommand\theequation{S\arabic{equation}}
\setcounter{figure}{0}
\setcounter{table}{0}
\setcounter{equation}{0}
\setcounter{page}{1}
\maketitlesupplementary

\section*{Appendix}

\section{Details on Evaluation metrics}
\label{sec:app_metrics}

We evaluate the models across five key dimensions: \textbf{aesthetic quality}, \textbf{semantic consistency}, \textbf{intra-shot consistency} (capturing subject and background stability), \textbf{inter-shot consistency}, and \textbf{transition control}.

\subsection{Transition Control Evaluation Metrics}
\label{subssec:app_SCAdefinition}

Furthermore, to comprehensively evaluate the model's ability to  to follow explicit shot-cut instructions, we propose the Shot Cut Accuracy (SCA) metrics,specified in appendix A. This metric holistically assesses shot control by quantifying both the accuracy of the number of cuts and the temporal precision of their placement. To compute the SCA, we first apply a pre-trained shot boundary detector, TransNet V2~\cite{transnetV2}, to the generated video to obtain the set of cut locations in the generated videos. These are then compared against the ground truth cut locations specified in the input. SCA is defined as:
\begin{equation}
    \text{SCA} = \exp(-\text{NSD})
\end{equation}
where NSD is the Normalized Shot Discrepancy, representing the total error relative to the video's length in frames, $F_{\text{total}}$:
\begin{equation}
    \text{NSD} = \frac{E_{\text{matched}}+E_{\text{penalty}}}{F_{\text{total}}}
\end{equation}
$E_{\text{matched}}$ quantifies the frame-wise deviation between predicted and ground-truth cuts after a one-to-one matching process.  $E_{\text{penalty}}$ is a penalty term for any missed or extraneous cuts. This ensures that models are penalized not only for imprecise timing but also for failing to produce the correct number of shots. The SCA score ranges in $(0, 1]$, where 1 indicates a perfect match. The exponential formulation makes the metric particularly sensitive to large errors, heavily penalizing significant temporal deviations.








\subsection{Aesthetic Quality}

We assess the aesthetic and artistic value of each video frame using the LAION aesthetic predictor~\cite{aesthetic}. This metric reflects human-perceived qualities such as composition, color harmony, realism, naturalness, and overall artistic appeal of the generated frames.

\subsection{Semantic Consistency.}
We evaluate the alignment between the text prompt and the generated video by measuring two types of semantic consistency: global and shot-level. For global consistency, we extract the representations of the entire prompt and the full video using ViCLIP~\cite{viclip} and compute their cosine similarity. For shot-level consistency, the video is divided into segments based on the input shot prompts, and the cosine similarity between each shot clip and its corresponding shot-level prompt features is calculated using ViCLIP.

\subsection{Intra-shot Consistency}

To compute intra-shot consistency, we first employ the pre-trained shot boundary detector TransNet V2~\cite{transnetV2} to identify cut locations within the generated videos. We then compute subject consistency and background consistency, following the design of VBench~\cite{vbench}. 

\noindent\textbf{Subject Consistency.}
For the main subject in the video, we measure the stability of its visual appearance across frames. Specifically, we extract DINO~\cite{dino} features for each frame and compute the average cosine similarity between consecutive frames and between each frame and the first frame.

\noindent\textbf{Background Consistency.}
To evaluate the temporal stability of the scene background, we compute CLIP~\cite{clip} feature similarities across frames. A higher similarity indicates a smoother and more coherent background transition over time.

\subsection{Inter-shot Consistency}

To assess consistency across different shots, a naive approach would be to extract ViCLIP features for each shot and compute the cosine similarity between them. However, since different shots may depict distinct characters or scenes, this simple comparison ignores diversity and may lead to biased results. To address this, we identify the characters described in the prompt and group the corresponding shots by character identity. We then compute the ViCLIP feature similarity among shots belonging to the same character group to obtain the inter-shot consistency score.